\documentclass[letterpaper]{article}

\usepackage{natbib,alifeconf}  
\usepackage{booktabs}

\usepackage{graphicx}
\usepackage{amsmath}
\usepackage{url,hyperref,cleveref}

\usepackage{amssymb}
\usepackage{booktabs}
\usepackage{multirow}
\usepackage{multicol}
\usepackage[table]{xcolor}

\usepackage{array}

\usepackage{pifont}
\definecolor{limegreen}{HTML}{32CD32}

\definecolor{pinkl}{HTML}{FB2B97}

\title{Emergent Dynamics in Neural Cellular Automata}

\author{
    Yitao Xu,
    Ehsan Pajouheshgar,
    Sabine Süsstrunk \\
    \mbox{}\\
    School of Computer and Communication Sciences, EPFL, Switzerland \\
    {\tt\small \{yitao.xu, ehsan.pajouheshgar, sabine.susstrunk \}@epfl.ch}
}

\begin{document}

\maketitle

\begin{abstract}
    Neural Cellular Automata (NCA) models are trainable variations of traditional Cellular Automata (CA). Emergent motion in the patterns created by NCA has been successfully applied to synthesize dynamic textures. However, the conditions required for an NCA to display dynamic patterns remain unexplored. Here, we investigate the relationship between the NCA architecture and the emergent dynamics of the trained models. Specifically, we vary the number of channels in the cell state and the number of hidden neurons in the MultiLayer Perceptron (MLP), and draw a relationship between the combination of these two variables and the motion strength between successive frames. Our analysis reveals that the disparity and proportionality between these two variables have a strong correlation with the emergent dynamics in the NCA output. We thus propose a design principle for creating dynamic NCA.
\end{abstract}

\section{Introduction}

Cellular Automata's (CA) ability to create complex patterns has fascinated researchers for many years \citep{von-ca, ca-conway, ca2003}. Neural Cellular Automata (NCA\footnote{We use NCA to refer to both Neural Cellular \textbf{Automata} and Neural Cellular \textbf{Automaton}.}), a more recent development \citep{mordvintsev2020growing}, introduces a trainable variant of conventional CA.
The emergence of complex and life-like behaviors, such as spontaneous motion and self-regeneration from simple local update rules, makes NCA a compelling candidate for studying artificial life.
NCAs have been applied to a variety of problems, including modeling morphogenesis \citep{mordvintsev2020growing, sudhakaran2021growing, sudhakaran2022goal}, texture synthesis \citep{niklasson2021self-sothtml, dynca, meshnca}, and collective modeling of intelligent behavior \citep{randazzo2020self-classifying, nca-collective-control}, demonstrating the effectiveness of NCAs in modeling complex biological and natural processes.

Recent studies \citep{dynca, meshnca} have particularly focused on utilizing emergent NCA dynamics to create moving patterns.  \cite{dynca} highlight that in NCA models trained for texture synthesis, the collective of cells exhibits spontaneously moving patterns. They propose a training schema to give structure to this random emergent motion and create dynamic textures. 
However, understanding the necessary conditions that give rise to such a spontaneous motion remains unexplored. Here, we investigate the relationship between emergent NCA dynamics and its hyperparameters, including the number of channels in the cell states ${C}$, and the number of hidden neurons $D$, in the MLP. We provide qualitative and quantitative results to show that NCA loses its dynamics as the disparity or ratio of $D$ and $C$ decreases. Based on our findings, we propose an NCA design principle to retain the dynamic properties.

\section{Preliminaries}

\subsection{Neural Cellular Automata}

\begin{figure}[!htbp]
    \centering
    \includegraphics[width=\linewidth,trim={200 235 220 210},clip]{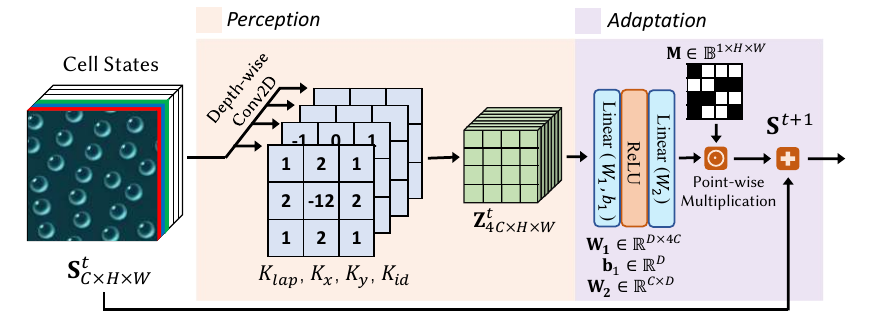}
    \caption{A single update step of NCA. }
    \vspace{-10pt}
    \label{fig:nca}
\end{figure}

In NCA, cells exist on a 2D grid, where each cell's state is represented by a $C$ dimensional vector. We refer to $C$ as the \textbf{Channel}. Given a 2D grid of size $H \times W$, all cells collectively define the state of the system $\mathbf{S} \in  \mathbb{R}^{C \times H \times W}$. An NCA iteratively updates cell states $\mathbf{S}$ over time by alternating between \textit{Perception} and \textit{Adaptation} stages. A single update step is visualized in Figure~\ref{fig:nca}.
In the \textit{Perception} stage, cells perceive their local 9-point Moore neighborhood via fixed convolution kernels. The typical choice \citep{mordvintsev2020growing, niklasson2021self-sothtml, dynca} of the convolution kernels are Sobel filters in the $x$ and $y$ directions, $K_x, K_y$, and a Laplacian filter $K_{lap}$.
Moreover, the cell itself is also counted as part of its neighborhood through convolution using an identity filter $K_{id}$. The output of the \textit{Perception} stage is then fed into an MLP, denoted $F_{\theta}$, where $\theta$ denotes the set of trainable parameters in the \textit{Adaptation} stage. $D$, also known as \textbf{Hidden}, is the number of hidden neurons in the MLP. Finally, the output of the MLP is multiplied by a stochastic binary mask, $\mathbf{M} \sim Bernoulli(0.5)$, to ensure asynchronous cell update, as this property is important for NCAs to behave well \citep{asynchronicity}. In summary, the NCA update rule can be expressed as:
\vspace{-3.5pt}
\begin{equation}
    \mathbf{S}^{t+1} = \mathbf{S}^{t} + F_{\theta}(\mathbf{S}^{t}, (K_x ,K_y , K_{lap}) * \mathbf{S}^{t}) \odot \mathbf{M}
\end{equation}
\subsection{Motion Quantification}
\label{sec:motion-quant}
The patterns created by the consecutive application of the NCA update rule display emergent and spontaneous motion. 
To quantify the strength of this motion, we utilize the pre-trained Optical Flow estimation network from \citep{two_stream}.
Given two consecutive frames from the sequence of patterns synthesized by an NCA, let $O_{ij} \in  \mathbb{R}^{2}$ be the estimated optical flow at the pixel $i, j$.  We define the motion strength $\Psi$ as the norm of optical flow averaged across all pixels:
\begin{equation}
    \Psi = \frac{1}{HW} \sum^{H}_{i=1} \sum^{W}_{j=1} ||O_{ij}||_2.
\end{equation}

\begin{figure}
    \centering
    \includegraphics[width=0.99\linewidth]{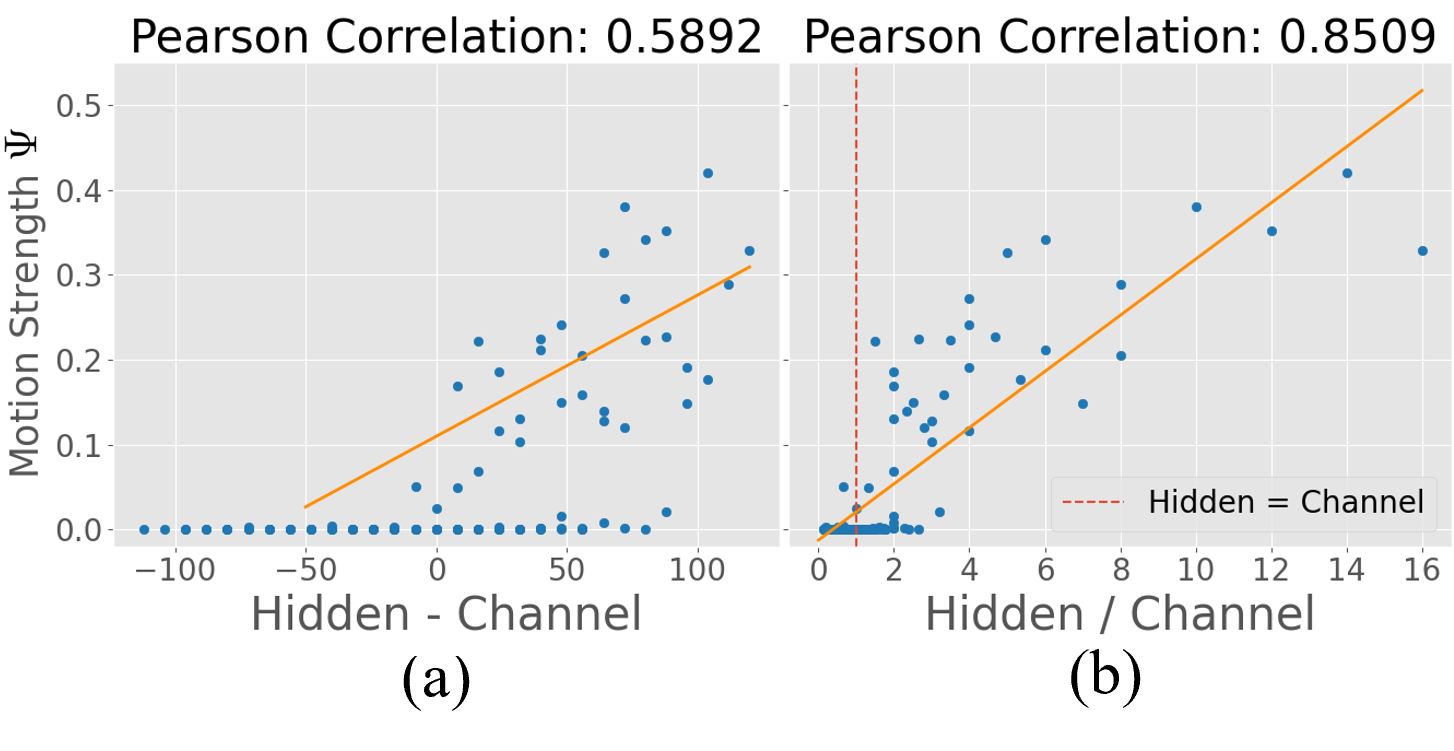}
    \vspace{-15pt}
    \caption{$Y$ axis shows the average motion strength $(\Psi)$. \textbf{(a):} $X$ axis shows Hidden - Channel $(D-C)$. Correlation significance: $p<0.002$. Small $Y$ values are excluded in the correlation fitting for a better visualization. With all values: $r=0.6225, p<0.0001$. \textbf{(b):} $X$ axis shows Hidden / Channel $(\frac{D}{C})$. Correlation significance: $p<0.0001$.}
    \label{fig:quant}
\end{figure}

\begin{figure}
    \centering
    \includegraphics[width=\linewidth,trim={180 200 200 180},clip]{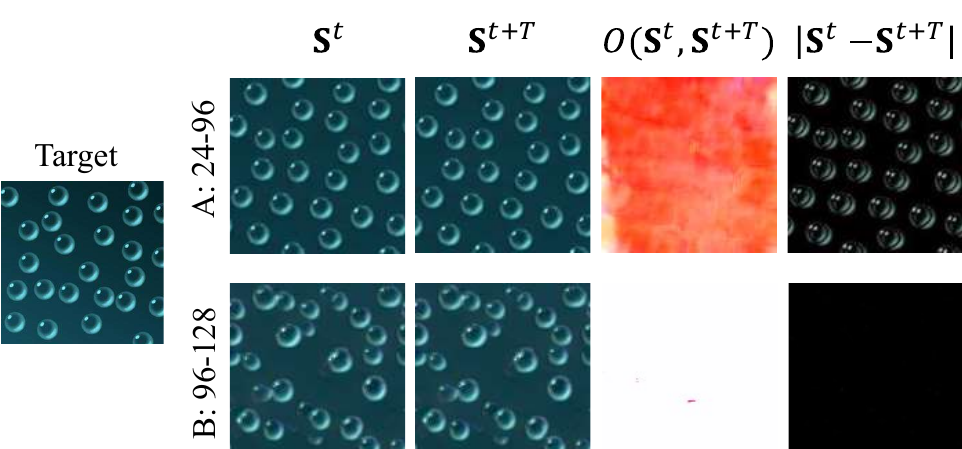}
    \caption{Visualization of frames during the test of NCA using two different numbers of channels $C$. Top row: $C=24,D=96$. Bottom row: $C=96, D=128$. $O(\mathbf{S}^t, \mathbf{S}^{t+T})$: the optic flow \citep{flow_visualization} between the two frames. $|\mathbf{S}^t-\mathbf{S}^{t+T}|$: the absolute difference between the two frames. Better viewed in \textcolor{pinkl}{\href{https://github.com/IVRL/Emergent_Dynamics_NCA/raw/main/video.mp4}{video}} format (download to local).}
    \label{fig:quali}
\end{figure}

\section{Experiments and Results}
We choose $C$ from \verb|range(8,128,8)|, and $D$ from \verb|range(16,128,16)|, where \verb|range(x,y,k)| means integer values from $x$ to $y$ inclusively with the step being $k$. We train NCAs on four different textures in the DTD dataset \citep{dtd}: \textit{bubbly\_0101, chequered\_0121, interlaced\_0172, cracked\_0085}, resulting in a total of 512 trained models. 
All models are trained for 6000 epochs to ensure convergence. We average the motion strength $\Psi$ across 100 frames for each NCA configuration. Each frame corresponds to $T$ NCA steps. We find that $T=32$ creates
sufficiently close yet distinct enough frames for accurate optic flow prediction. A smaller $T$ would result in minimal motion that makes it challenging for the Optic Flow network to detect, while a larger $T$ produces excessive displacement, leading to incorrect optic flow estimations.
Moreover, we observe that all NCAs become stable after around 100 frames. Therefore,  we first let the NCA generate 100 frames and then perform the measurement on the second 100 frames.
Figure~\ref{fig:quant}-(a) shows the relationship between motion strength and $D - C$ for all NCA configurations. 
We observe that once $D < C$, the emergent motion completely disappears and as the difference $D-C$ increases, the NCA becomes more dynamic. However, there are still some NCA models with $D>C$ that are not dynamic. We further investigate this problem by plotting the ratio instead of the difference between $D$ and $C$ in Figure~\ref{fig:quant}-(b). We notice that NCA barely has motion when $\frac{D}{C}<2$, which accounts for the case when there is no motion, but $D>C$. We visualize the output of two NCA models, A: $C=24,D=96$ and B: $C=96,D=128$, in Figure~\ref{fig:quali} to demonstrate an example of dynamic and non-dynamic NCA configurations. Although NCA B satisfies $D>C$, the ratio between $D$ and $C$ is too small for the model to permit dynamic patterns.

\section{Discussion and Conclusion}
We empirically find that the number of hidden neurons should be larger than the number of channels and that their ratio should be at least greater than 2.0 for an NCA to display emergent dynamics. 
Without the dynamic properties, NCA is prone to get stuck in local minima where artifacts appear in the generated texture, as shown in Figure~\ref{fig:quali}. Moreover, the lack of dynamics will render NCA no longer suitable for certain applications, such as dynamic texture synthesis and biological pattern simulation. Therefore, in practice, if we favor dynamic NCAs without artifacts, it is critical to ensure the \textbf{Hidden} $D$ is larger by a factor of two or more than the \textbf{Channel} $C$.

\clearpage

\footnotesize
\bibliographystyle{apalike}
\bibliography{main} 

\begin{thebibliography}{}

\bibitem[Baker et~al., 2011]{flow_visualization}
Baker, S., Scharstein, D., Lewis, J., Roth, S., Black, M.~J., and Szeliski, R. (2011).
\newblock A database and evaluation methodology for optical flow.
\newblock {\em International Journal of Computer Vision}, 92(1):1--31.

\bibitem[Cimpoi et~al., 2014]{dtd}
Cimpoi, M., Maji, S., Kokkinos, I., Mohamed, S., , and Vedaldi, A. (2014).
\newblock Describing textures in the wild.
\newblock In {\em Proceedings of the {IEEE} Conf. on Computer Vision and Pattern Recognition ({CVPR})}.

\bibitem[Gardner, 1970]{ca-conway}
Gardner, M. (1970).
\newblock Mathematical games.
\newblock {\em Scientific american}, 222(6):132--140.

\bibitem[Mordvintsev et~al., 2020]{mordvintsev2020growing}
Mordvintsev, A., Randazzo, E., Niklasson, E., and Levin, M. (2020).
\newblock Growing neural cellular automata.
\newblock {\em Distill}.
\newblock https://distill.pub/2020/growing-ca.

\bibitem[Nadizar et~al., 2022]{nca-collective-control}
Nadizar, G., Medvet, E., Nichele, S., and Pontes-Filho, S. (2022).
\newblock Collective control of modular soft robots via embodied spiking neural cellular automata.
\newblock {\em arXiv preprint arXiv:2204.02099}.

\bibitem[Niklasson et~al., 2021a]{asynchronicity}
Niklasson, E., Mordvintsev, A., and Randazzo, E. (2021a).
\newblock Asynchronicity in neural cellular automata.
\newblock In {\em Artificial Life Conference Proceedings 33}, volume 2021, page 116. MIT Press One Rogers Street, Cambridge, MA 02142-1209.

\bibitem[Niklasson et~al., 2021b]{niklasson2021self-sothtml}
Niklasson, E., Mordvintsev, A., Randazzo, E., and Levin, M. (2021b).
\newblock Self-organising textures.
\newblock {\em Distill}, 6(2):e00027--003.

\bibitem[Pajouheshgar et~al., 2024]{meshnca}
Pajouheshgar, E., Xu, Y., Mordvintsev, A., Niklasson, E., Zhang, T., and S{\"u}sstrunk, S. (2024).
\newblock Mesh neural cellular automata.
\newblock {\em ACM Trans. Graph.}

\bibitem[Pajouheshgar et~al., 2023]{dynca}
Pajouheshgar, E., Xu, Y., Zhang, T., and S\"usstrunk, S. (2023).
\newblock Dynca: Real-time dynamic texture synthesis using neural cellular automata.
\newblock In {\em Proceedings of the IEEE/CVF Conference on Computer Vision and Pattern Recognition (CVPR)}, pages 20742--20751.

\bibitem[Randazzo et~al., 2020]{randazzo2020self-classifying}
Randazzo, E., Mordvintsev, A., Niklasson, E., Levin, M., and Greydanus, S. (2020).
\newblock Self-classifying mnist digits.
\newblock {\em Distill}.
\newblock https://distill.pub/2020/selforg/mnist.

\bibitem[Sudhakaran et~al., 2021]{sudhakaran2021growing}
Sudhakaran, S., Grbic, D., Li, S., Katona, A., Najarro, E., Glanois, C., and Risi, S. (2021).
\newblock Growing 3d artefacts and functional machines with neural cellular automata.
\newblock In {\em Artificial Life Conference Proceedings 33}, volume 2021, page 108. MIT Press One Rogers Street, Cambridge, MA 02142-1209.

\bibitem[Sudhakaran et~al., 2022]{sudhakaran2022goal}
Sudhakaran, S., Najarro, E., and Risi, S. (2022).
\newblock Goal-guided neural cellular automata: Learning to control self-organising systems.
\newblock {\em arXiv preprint arXiv:2205.06806}.

\bibitem[Tesfaldet et~al., 2018]{two_stream}
Tesfaldet, M., Brubaker, M.~A., and Derpanis, K.~G. (2018).
\newblock Two-stream convolutional networks for dynamic texture synthesis.
\newblock In {\em Proceedings of the IEEE Conference on Computer Vision and Pattern Recognition}, pages 6703--6712.

\bibitem[Von~Neumann et~al., 1966]{von-ca}
Von~Neumann, J., Burks, A.~W., et~al. (1966).
\newblock Theory of self-reproducing automata.
\newblock {\em IEEE Transactions on Neural Networks}, 5(1):3--14.

\bibitem[Wolfram and Gad-el Hak, 2003]{ca2003}
Wolfram, S. and Gad-el Hak, M. (2003).
\newblock A new kind of science.
\newblock {\em Appl. Mech. Rev.}, 56(2):B18--B19.

\end{thebibliography}

\end{document}